\pdfoutput=1

\PassOptionsToPackage{capitalize}{cleveref}
\documentclass[11pt]{article}

\usepackage{ACL}
\usepackage[times]{stroop}
\let\cref\Cref

\usepackage{times}
\usepackage{latexsym}

\usepackage[T1]{fontenc}

\usepackage[utf8]{inputenc}

\usepackage{microtype}

\usepackage{inconsolata}
\usepackage{amsmath}
\usepackage{graphicx}
\usepackage{caption}
\usepackage[list=true]{subcaption}
\usepackage{graphbox}
\usepackage{pgfplots}
\usepackage{tikzscale}
\usepackage{float}
\usepackage{tabularx}

\title{On Measuring Context Utilization in Document-Level MT Systems}

\author{Wafaa Mohammed  \qquad Vlad Niculae  \\
  Language Technology Lab \\ University of Amsterdam \\
  \texttt{\{w.m.a.mohammed, v.niculae\}@uva.nl}
}

\begin{document}
\maketitle

\begin{abstract}
Document-level translation models are usually evaluated
using general metrics such as BLEU, which are not informative about the benefits of context. Current work on context-aware evaluation, such as contrastive methods, only measure translation accuracy
on words that need context for disambiguation. Such measures cannot reveal whether the translation model uses the correct supporting context. We propose to complement accuracy-based evaluation with measures of context utilization. We find that perturbation-based analysis (comparing models' performance when provided with correct versus random context) is an effective measure of overall context utilization. For a finer-grained phenomenon-specific evaluation, we propose to measure how much the supporting context contributes to handling context-dependent discourse phenomena. We show that automatically-annotated supporting context gives similar conclusions to human-annotated context and can be used as alternative for cases where human annotations are not available. Finally, we highlight the importance of using discourse-rich datasets when assessing context utilization.
\footnote{Code at \href{https://github.com/Wafaa014/context-utilization}{https://github.com/Wafaa014/context-utilization}.}
\end{abstract}

\section{Introduction}

Documents are one of the primary ways in which we produce and consume text.
While for some languages, sentences provide a base unit of meaning, there are many sentences that contain discourse phenomena that are difficult to disambiguate at sentence level (\cref{example}). Despite the vital need for document-level translation in order to handle context-dependent phenomena, most of the current works on machine translation focus on sentence-level translation. \Citet{DBLP:journals/corr/abs-2304-12959} listed the problem of evaluation as one of the reasons for the inability to move beyond sentence level.
In this work, we focus on this problem of evaluation. In particular, we focus on evaluating document-level translation models based on how well they utilize inter-sentential information provided when translating at the document level.

The research on document-level translation evaluation has progressed significantly. Early works used general metrics such as BLEU \citep{DBLP:conf/acl/PapineniRWZ02} and TER \citep{DBLP:conf/amta/SnoverDSMM06} which proved to be inadequate for capturing improvements in discourse phenomena. Subsequent research introduced phenomena-specific automatic metrics and contrastive test suites. \Citeposs{DBLP:journals/csur/MarufSH21} survey includes a comprehensive list of works in this direction. While these metrics provide an accuracy measure of models' performance on phenomena, they do not account for correct context utilization.
Unlike prior studies, we adopt an interpretable approach to context utilization evaluation. We evaluate models based on the ability to use the correct context, and not only the ability to generate a correct translation without necessarily utilizing the context.

To assess models' correct context utilization, we perform a perturbation-based analysis. Previous studies in perturbation analysis, such as the works of \citet{DBLP:conf/acl/VoitaST20}, \citet{DBLP:conf/acl/LiLWJXZLL20}, and \citet{DBLP:journals/corr/abs-2109-02995}, were limited to specific architectures, evaluated on particular metrics, and perturbed only the source context. In a more comprehensive study, we analyze performance across various document-level architectures using multiple metrics: BLEU, COMET \citep{DBLP:conf/wmt/ReiSAZFGLCM22} and CXMI \citep{DBLP:conf/acl/FernandesYNM20}. Additionally, our analysis involves perturbing both source and target contexts to examine the influence of both sides.

For more fine-grained analysis at the level of a specific discourse phenomenon, \citet{DBLP:conf/acl/YinFPCMN20}  collected annotations of supporting context words from expert translators for the pronoun resolution phenomenon. They propose using such annotations as supervision to guide models' attention. Extending their work, we focus on benchmarking context-aware models' performance on the phenomenon. We evaluate models based on the attribution scores of supporting context. To obtain attribution scores, we use one of the state-of-the-art interpretability methods for transformer models: ALTI+ \citep{DBLP:conf/emnlp/FerrandoGAEC22}. Moreover, we use automatically annotated (using coreference resolution models) supporting context as an alternative to human annotated context and show that it gives similar conclusions. Using automatic annotations allowed us to scale to different languages and has the  potential to extend to other discourse phenomena.

As an accuracy measure on discourse phenomena, \citet{DBLP:conf/acl/FernandesYLMN23} proposed a novel systematic approach to tag words in a corpus with specific discourse phenomena and evaluate models' performance using F1 measure. However, they mention that context-aware models make only marginal improvements over context-agnostic models. Our analysis reveals that this depends on the richness of the dataset with phenomena, and that challenge sets curated to target context-dependent discourse phenomena are better in distinguishing the differences between models in handling the phenomena.

Our contributions are the following:
\begin{itemize}
    \item We perform a perturbation-based analysis on document-level models and find that single-encoder concatenation models are able to make use of the correct context vs.\ a random context.

    \item We propose the use of attribution scores of \textit{supporting context} to evaluate correct context utilization. Analyzing the pronoun resolution phenomenon as a
    case study, we find that sentence-level models and single-encoder context-aware models are better than multi-encoder models in terms of the amount of attribution pronoun's antecedents have to generating the pronoun.

    \item We propose the use of automatically annotated \textit{supporting context} as an alternative to human-annotated context for attribution evaluation. We show that, despite noise in automatic annotation, results are consistent with human-annotated context, paving the way towards efficient use of linguistic expertise in document-level translation evaluation.

    \item We highlight the importance of using a discourse rich dataset when evaluating the ability of models to handle context-dependent discourse phenomena.

\end{itemize}

\begin{figure}\footnotesize
\hrule
[EN] One of the Chinese worked in an \colorbox{cyan!20}{\underline{amusement park}}.
\textbf{It} was closed for the season.\\
\break
[DE] Ein Chinese arbeitete in einem \colorbox{cyan!20}{\underline{Vergnügungspark}}.
\textbf{Er} war gerade geschlossen. \\
\hrule
\caption{An example illustrating the pronoun resolution phenomena which can not be disambiguated at sentence level. The pronoun \textbf{It} is ambiguous and its translation depends on the \colorbox{cyan!20}{\underline{antecedent}}.\footnotemark}

\label{example}
\end{figure}

\footnotetext{Example is drawn from ContraPro dataset \href{https://github.com/ZurichNLP/ContraPro}{https://github.com/ZurichNLP/ContraPro}}

\section{Background}
Sentence-level MT models treat sentences in a document as separate units. They only consider intra-sentential dependencies. In contrast, document-level models take into account intra-sentential as well as inter-sentential dependencies. Formally, if we consider a document
containing parallel sentences
$D = \{(x_1,y_1), (x_2,y_2),... , (x_n, y_n)\}$, the probability of translating sentence $x_i$ into $y_i$ using a sentence-level model is
\[ P(y_i | x_i) = \prod_{t=1}^{T} P (y_{i,t} | y_{i,<t}, x_i), \]
while the probability using a document-level translation model with context $C_i$ is:
\[ P(y_i | x_i, C_i) = \prod_{t=1}^{T_i} P (y_{i,t} | y_{i,<t}, x_i, C_i), \]
where $T_i$ is the 
token length of sentence $y_i$,
and \(C_i\) may contain source and target context, as desired.

There are several ways to design neural architectures for document-level MT. The main architectures developed so far can be broadly classified into two categories based on how they combine the context and current sentence representations: single-encoder and multi-encoder approaches.

\subsection{Single-Encoder Approaches}
\label{concatenation}
The single-encoder approach to document level MT works by concatenating previous sentences to the current sentence separated by a special token. It is commonly deployed under two setups: a 2-to-2 setup in which the previous and current source sentences are translated together, the translation of the current source sentence is then obtained by extracting tokens after the special concatenation token on the target side, and a 2-to-1 setup where the concatenation happens only in the source side, the target in this case is only the current sentence translation \citep{DBLP:conf/discomt/TiedemannS17,DBLP:conf/naacl/BawdenSBH18}.

\subsection{Multi-Encoder Approaches}
\label{multi-encoder-approaches}
The multi-encoder approach uses extra encoders for source and target contexts. The encoded representations of the context and current sentences are combined together before being passed to the decoder. There are different ways to combine the context and current sentence representations. Methods in the literature include concatenation, hierarchical attention, and attention gating \citep{DBLP:conf/acl/LibovickyH17,DBLP:conf/naacl/ZophK16,DBLP:conf/emnlp/WangTWL17,DBLP:conf/naacl/BawdenSBH18}.

\section{Experimental details}

\subsection{Data}

 We train our models on IWSLT2017 TED data \citep{DBLP:conf/eamt/CettoloGF12}. We consider two language pairs in our experiments, namely EN $\rightarrow$ DE and EN $\rightarrow$ FR.  For EN $\rightarrow$ DE, we use the same splits used by \citet{DBLP:conf/naacl/MarufMH19};  we combine \texttt{tst2016--2017} into the test set and the rest are used for development. For EN $\rightarrow$ FR, we use the same splits as \citet{DBLP:conf/acl/FernandesYNM20}; we use the sets \texttt{tst2011--2014} 
 as validation sets and \texttt{tst2015} as the test set.

\subsection{Models}

\begin{figure*}
\begin{subfigure}[c]{0.3\textwidth}
\includegraphics[width=5cm]{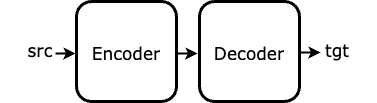}
\caption{Sentence-level Model}
\label{sentence_model}
\end{subfigure}\hfill
\begin{subfigure}[c]{0.3\textwidth}
\includegraphics[width=5cm]{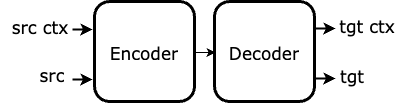}
\caption{Concatenation Model}
\label{concat_model}
\end{subfigure}\hfill
\begin{subfigure}[c]{0.3\textwidth}
\includegraphics[width=5cm]{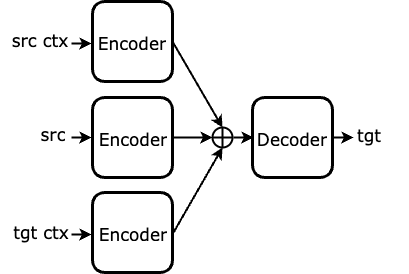}
\caption{Multi-encoder Model}
\label{multi_encoder_model}
\end{subfigure}
\caption{Model architectures for different settings. src \& tgt refer to the current source and target sentence pair. src ctx \& tgt ctx refer to the previous source and target sentence pairs used as context. In the concatenation model, the context and current sentences are concatenated together with a special separator token in between them. In the multi-encoder model, the $\oplus$ symbol refers to a concatenation operation.}
\end{figure*}

 For both language pairs, we consider an encoder-decoder transformer architecture as our base model \citep{DBLP:conf/nips/VaswaniSPUJGKP17}. Similar to \citet{DBLP:conf/acl/FernandesYNM20}, we train a transformer small model (hidden size of 512, feedforward size of 1024, 6 layers, 8 attention heads). All models are trained on top of Fairseq \citep{DBLP:conf/naacl/OttEBFGNGA19}. We use the same hyper-parameters as \citet{DBLP:conf/acl/FernandesYNM20}, we train using the Adam optimizer with $\beta_1 = 0.9$ and $\beta_2 = 0.98$ and use an inverse square root learning rate scheduler with an initial value of $5 \times 10^{-4}$ and with a linear warm-up in the first 4000 steps.
We train the models with early stopping on the validation perplexity. For models that use context, we train the models using a dynamic context size of 0--5 previous source and target sentences to ensure robustness against varying context size, as recommended by \citet{DBLP:conf/acl/SunWZZHCL22}.
We develop three models for our evaluation experiments:
 \begin{itemize}
     \item \textbf{A sentence-level model:} As in \cref{sentence_model}, we train an encoder-decoder model on sentence-level data. This model has two evaluation setups: a sentence-level and a document-level setup. When evaluating at the sentence level, we refer to this model as the \textbf{sentence-level (sent)} model. To perform document-level evaluation, context and current sentences are concatenated with a special separator token in between them;
we refer to this scenario as the \textbf{sentence-level*} model.

     \item \textbf{A single-encoder concatenation model:} As seen in \cref{concat_model}, we use the 2-to-2 setup (\cref{concatenation}) with a sliding window across sentences in each document, allowing us to consider both source and target contexts. We refer to this model as the \textbf{concatenation} model.

     \item \textbf{A multi-encoder concatenation model:} As in \cref{multi_encoder_model}, we add two extra encoders to represent source and target contexts. The outputs of the three encoders are concatenated before being passed to the decoder. We refer to this model as the \textbf{multi-encoder} model. Per \cref{multi-encoder-approaches}, there are other methods to combine the outputs of multiple encoders beyond concatenation. However, we opt for concatenation due to its simplicity and its comparable BLEU performance to other architectures, as presented in \citet{DBLP:conf/naacl/BawdenSBH18}.
 \end{itemize}

\section{Method}
Our goal is to build interpretable metrics to measure the extent of context utilization in context-aware MT.
To this end, we propose two methods: a perturbation analysis and an attribution analysis.

\subsection{Perturbation-Based Analysis}
We look at the difference in performance when passing the correct versus random tokens as context. The correct context is the previous 5 sentences on source side, and the previous 5 generated translations on the target side.\footnote{We avoid using the gold target context at inference time to eliminate exposure bias.} To generate random context, we sample random tokens from the model's vocabulary with a size similar to the correct context size. We compare models across BLEU, COMET and CXMI \citep[conditional cross-mutual information,][]{DBLP:conf/acl/FernandesYNM20} metrics. CXMI is used to measure context usage by comparing the model distributions over a dataset with and without context. It should be noted that the numerical CXMI value cannot be compared across models since the multi-encoder model has a different number of parameters which will affect the probability distribution learned by the model. Therefore, we mainly focus on the sign of the CXMI value for the comparison. A positive CXMI value means that introducing context increases the probabilities assigned by the model to output tokens, and a negative CXMI means that the context is reducing them.  Formally, for a source--target pair $(x,y)$ and a context $C$,
it reads:
\begin{multline*}
     \operatorname{CXMI}(C \rightarrow y |x) = \\
     H_{q_{\text{MT}_{A}}} (y |x) - H_{q_{\text{MT}_{C}}} (y |x, C),
\end{multline*}
where $H_{q_{MT_{A}}}$ is the entropy of the context agnostic model and $H_{q_{MT_{C}}}$ is the entropy of the context-aware model.
In our analysis, we evaluate the same model with and without context, i.e., $q_{\text{MT}_{A}} = q_{\text{MT}_{C}} = q_{\text{MT}}$

We compute the BLEU score using sacreBLEU \citep{post-2018-call,DBLP:conf/acl/PapineniRWZ02} and the
COMET score
\citep{rei-etal-2020-comet,rei-etal-2022-comet} using the \textit{wmt22-comet-da} model\footnote{\href{https://huggingface.co/Unbabel/wmt22-comet-da}{https://huggingface.co/Unbabel/wmt22-comet-da}} and directly compare the numerical values of the scores in the correct vs. random context setup.
Besides the high BLEU and COMET performance under the correct context setup, we regard models that show a difference in performance between the correct and random context setups as utilizing the correct context.

\subsection{Attribution Analysis}
In this experiment, we measure the attribution of supporting context words to model predictions. By \textit{supporting context words}, we mean the words that are necessary to resolve context-dependent phenomena. For example, in case of pronoun resolution, the supporting context words are the pronoun's antecedents.

We look at the percentage of attribution of pronoun antecedents to generating a pronoun against the attribution of the entire input. We make use of the ContraPro contrastive evaluation dataset for the analysis. For EN $\rightarrow$ DE, the dataset considers the translation of the English pronoun $it$ to the three German pronouns \textit{er}, \textit{sie} or \textit{es}. It consists of 4K examples per pronoun \citep{DBLP:conf/wmt/MullerRVS18}. For EN $\rightarrow$ FR, the dataset concerns the translation of the English pronouns \textit{it}, \textit{they} to their French correspondents \textit{il}, \textit{elle}, \textit{ils}, and \textit{elles}. It includes 14K samples evenly split across the pronouns \citep{DBLP:conf/eamt/LopesFBZM20}. In particular, we use a subset of the data that has an antecedent distance between 1--5 since we are using 5 previous sentences as context.\footnote{For EN$\rightarrow$DE, we exclude 2400 examples with antecedent distance 0, and 118 examples with a distance greater than 5. For EN$\rightarrow$FR, 5986 examples with distance 0 are excluded.}

The attribution method we used is the ALTI+ (Aggregation of Layer-wise Token-to-token Interactions) method \citep{DBLP:conf/emnlp/FerrandoGAEC22}, which has been shown to be effective in explaining model behaviors \citep[e.g. detecting hallucinations,][]{DBLP:conf/acl/DaleVBC23}.
ALTI+ is an interpretability method used to track the attributions of input tokens (\textbf{source sentence} and \textbf{target prefix}) through an attention rollout method. In ALTI+, the information flow in the transformer model is treated as a directed acyclic graph and the amount of information flowing from one node to another in different layers is computed by summing over the different paths connecting both nodes, where each path is the result of the multiplication of every edge in the path.

\textbf{Source sentence} contributions are computed by the matrix multiplication of the layer-wise contributions, giving the full encoder contribution matrix $\mbC^{enc}_{e\leftarrow x}$. This can be readily applied for both the sentence-level and concatenation models. However, further consideration is needed to apply it in the multi-encoder setup. In the multi-encoder model, the input consists of separate source context, source, and target context sequences
$x = [x_{sc}, x_{s}, x_{tc}]$. Each sequence is encoded separately by a different encoder giving ALTI contribution matrices $\mbC^{enc_{sc}}_{e_{sc}\leftarrow x_{sc}}$, $\mbC^{enc_{s}}_{e_{s}\leftarrow x_{s}}$ and $\mbC^{enc_{tc}}_{e_{tc}\leftarrow x_{tc}}$, respectively. Since we concatenate the output of each encoder giving $e = [e_{sc}, e_{s}, e_{tc}]$, the overall encoder contribution matrix is block diagonal:
\[
\mbC^{enc}_{e\leftarrow x} =
\begin{bmatrix}
\mbC^{enc_{sc}}_{e_{sc}\leftarrow x_{sc}} & \bm{0} & \bm{0} \\
\bm{0} & \mbC^{enc_{s}}_{e_{s}\leftarrow x_{s}} & \bm{0} \\
\bm{0} & \bm{0} & \mbC^{enc_{tc}}_{e_{tc}\leftarrow x_{tc}} \\
\end{bmatrix}.
\]
The rest of the ALTI+ method proceeds unchanged, as explained in \cite[section 3]{DBLP:conf/emnlp/FerrandoGAEC22}. It includes multiplying each of the cross-attention contribution matrices with the contributions of the entire encoder to account for all the paths in the encoder. Afterwards, edges from paths of \textbf{target prefix} contributions are aggregated.

We obtain word-level attribution scores and then compute the percentage of the sum of attributions of source and target antecedent words against the total attribution of the entire input.\footnote{We compute the scores for the first occurrence of the antecedent. This might penalize a model that pays attention to another occurrence of the antecedent. This is rare: the average number of antecedents is 1.09 for DE and 1.18 for FR.}

\section{Results and Discussion}

\begin{table}
\centering
\begin{tabular}{l@{\!\!}r@{\hspace{.9em}}r@{\hspace{.9em}}r}
\hline
& \textbf{antecedents} & \textbf{context} & \textbf{current} \\
\hline
\textbf{ContraPro DE} \\
\hline
sentence-level & 0.00 & 0.00 & 100.00 \\
sentence-level* & 1.69& 89.71 & 10.29 \\
concatenation & 2.86 & 78.09 & 21.91 \\
multi-encoder & 0.07 & 2.36 & 97.64 \\
\hline
\textbf{ContraPro FR} \\
\hline
sentence-level &  0.00 & 0.00 & 100.00 \\
sentence-level* &  3.57 & 84.38 & 15.62\\
concatenation & 2.59 & 76.19 & 23.81\\
multi-encoder &  0.25 & 3.07 & 96.93\\
\hline
\end{tabular}
\caption{The percentage of attribution of pronouns' antecedents, the entire
context words, and current sentence words to generating the ambiguous pronoun
in the ContraPro dataset.}
\label{alti}
\end{table}

\subsection{Are Models Sensitive\linebreak To The Correct Context?}
\label{perturb}
Results of the perturbation analysis are shown in \cref{perturbaion}. For both language pairs, the concatenation model is making use of correct context tokens, and presenting random context tokens to the model results in worse BLEU and COMET performances and a negative CXMI value. Even though the sentence-level model has high BLEU and COMET scores, its performance drops significantly when evaluated at the document level (sentence-level*). This is expected; since the model has not been trained on longer contexts. Regarding the multi-encoder model, even though it has the best BLEU score for both language pairs and the best COMET score for EN$\rightarrow$DE, the consistent performance of the model with correct and random context suggests that it is not utilizing the correct context, consistent with the low or negative CXMI values. This analysis highlights the importance of looking beyond the BLEU and COMET scores when evaluating context utilization of document-level MT models.

\begin{table*}
\centering
\begin{tabularx}{.99\textwidth}{l@{\hspace{2em}} r r r @{\hspace{2.7em}} r r r
@{\hspace{2.7em}} r r}
\hline
& \multicolumn{3}{l}{\hspace{2.5em}\textbf{BLEU}} & \multicolumn{3}{l}{\hspace{2em}\textbf{COMET}} &  \multicolumn{2}{l}{\hspace{2.5em}\textbf{CXMI}} \\
\hline
\textbf{setup} & rand & correct & no-ctx & rand & correct & no-ctx & rand & correct \\
\hline
{\textbf{EN$\rightarrow$DE}} \\
\hline
sentence-level & -- & -- & 23.2 & -- & -- & 75.1   & -- & -- \\
sentence-level* & 2.5 & 3.5 & --  & 33.7  & 42.0 & --  & \colorbox{red!40}{$-2.980$} & \colorbox{red!40}{$-2.100$} \\
concatenation & 20.2 & 23.3 & 23.4  & 68.2 & 75.4 & 75.4  & \colorbox{red!40}{$-0.320$} & \colorbox{green!50}{+0.014}\\
multi-encoder & 23.7 & \textbf{23.7} & 23.7 &  75.7 & \textbf{75.8} & 75.9   & \colorbox{red!40}{$-0.002$} & \colorbox{red!40}{$-0.002$}\\
\hline
{\textbf{EN$\rightarrow$FR}} \\
\hline
sentence-level & -- &  -- & 36.2	& -- & -- & \textbf{78.2}   & --	& -- \\
sentence-level* &  5.6	& 9.4 & --  & 36.2 & 46.6 & --  & \colorbox{red!40}{$-2.950$}	& \colorbox{red!40}{$-1.840$}\\
concatenation & 27.9 &	35.6 &	35.8 & 65.8 & 77.6 & 77.8  & \colorbox{red!40}{$-0.320$} & \colorbox{green!50}{+0.006}\\
multi-encoder & 36.9 &	\textbf{36.9} & 36.6 & 77.9 & 77.9 & 78.0   & \colorbox{green!50}{+0.002}	& \colorbox{green!50}{+0.002}\\
\hline
\end{tabularx}
\caption{BLEU, COMET and CXMI scores of correct vs.\ random
context on IWSLT2017 test data. The best BLEU and COMET scores in a correct setup (with context for the concatenation and multi-encoder models and without context for the sentence-level model) are bolded.
High BLEU and COMET scores, as well as a difference in performance between the correct and random context setups are expected for effective context utilization, as demonstrated by the concat model.
A \colorbox{green!50}{positive} CXMI value means that the probabilities of output tokens are increased with context while a \colorbox{red!40}{negative} CXMI value means that context is reducing them.}
\label{perturbaion}
\end{table*}

\subsection{Are Models Paying ``Attention''\linebreak To The Supporting Context?}
We obtain the attribution scores of the \textit{supporting context} provided in the ContraPro pronoun resolution dataset. The \textit{supporting context} is automatically generated using coreference resolution tools. Looking at \cref{alti}, we can see that the sentence-level* model and the concatenation model have higher attribution scores compared to the multi-encoder model. This can also be confirmed by the low overall context attribution compared to the current sentence attribution in the multi-encoder model. It should be noted that our implementation of the multi-encoder model depends on simple concatenation of the encoders' outputs before being fed to the decoder. More complicated multi-encoder setups (\eg, using gating mechanisms or hierarchical attention) might have better context attribution. Moreover, for German pronouns, looking at the total context contributions, we observe that despite the fact that the sentence-level* model has the highest context attributions, it is not the best at utilizing the \textit{supporting context}. This highlights the importance of focusing on important parts of the context when evaluating context utilization.

\subsection{Does Automatically Annotated Supporting Context Align With Human Annotated Supporting Context?}
 We investigate whether the automatically annotated \textit{supporting context} aligns with the way humans utilize context for pronoun disambiguation. We use the SCAT (Supporting Context for Ambiguous Translations) data provided by \citet{DBLP:conf/acl/YinFPCMN20} which contains human annotations of \textit{supporting context} for pronoun resolution on the French ContraPro data. We filter the data for instances that has an antecedent outside the current sentence and end up with 5961 instances for evaluation. We calculate the attribution scores of human context for the models we built for EN$\rightarrow$FR translation. Comparing the attribution percentages in \cref{alti-scat} to the attributions on ContraPro FR data in \cref{alti}, we observe similar trends across models. The sentence-level* and concatenation models have comparable attribution scores and are higher than the multi-encoder model. This shows that automatically annotated context can be a good alternative to human annotations which are expensive to obtain at scale.


\begin{table}
\centering
\begin{tabular}{l@{\!\!}r@{\hspace{.9em}}r@{\hspace{.9em}}r}
\hline
\textbf{model} & \textbf{antecedents} & \textbf{context} & \textbf{current} \\
\hline
sentence-level & 0.00  & 0.00 & 100.00 \\
sentence-level* &  1.25 & 87.12 & 12.88\\
concatenation & 1.03 & 74.23 & 25.77\\
multi-encoder &  0.53 & 2.49 & 97.50\\
\hline
\end{tabular}
\caption{Attribution percentages of human annotated antecedents, the entire
context words, and current sentence words to generating the ambiguous pronoun
in the SCAT dataset.}
\label{alti-scat}
\end{table}

\subsection{Are Models Able To Handle Context-Dependent Phenomena?}
The ultimate goal of context-aware MT is being able to model context-dependent phenomena. Hence, we evaluate models on their ability to address these phenomena. We use the Multilingual Discourse Aware benchmark (MuDA) to automatically tag datasets with context-dependent phenomena \citep{DBLP:conf/acl/FernandesYLMN23}. We consider four linguistic discourse phenomena in our analysis: lexical cohesion, formality, pronoun resolution and verb form.
\textbf{Lexical cohesion} refers to consistently translating an entity in the same way throughout a document. 
\textbf{Formality} is the phenomenon where the second-person pronoun that the speaker uses depends on their relationship the the person being addressed. 
\textbf{Pronoun resolution} denotes the phenomenon in languages that use gendered pronouns for pronouns other than the third-person singular, or assign gender based on formal rules instead of semantic ones. 
\textbf{Verb form} denotes the phenomenon in languages with a fine-grained verb morphology, where the translation of the verb should reflect the tone, mood and cohesion of the document.

We use the IWSLT2017 test set as well as ContraPro data (including context sentences) in the analysis. \Cref{phenomena-statistics} presents the statistics of discourse phenomena in these datasets. We then evaluate models using the F1 measure based on whether a word tagged in the reference exists and is also tagged in the hypothesis.
As can be seen in \cref{phenomena-statistics}, for both language pairs, ContraPro dataset has a higher percentage of tokens tagged with pronouns (since the dataset targets this phenomena). Looking at the F1 measure of models on this dataset in \cref{f-measure-pronouns}, we can see that the concatenation model has a higher score compared to other models which is reflected in the ContraPro accuracy as well (\cref{ContraPro-accuracy}). On the other hand, the lower percentages of phenomena in the IWSLT data results in similar performance across models on this data. We highlight the importance of using a discourse rich dataset to benchmark models' performance on handling context-dependent phenomena. Evaluation on other discourse phenomena, which neither of the datasets targeted, resulted in no distinction between the models as seen in \cref{f-measure-other-phenomena,f-measure-fr}.  The low F1 measure of the sentence-level* model across phenomena on the IWSLT data can be linked to its low translation performance as presented in \cref{perturb}. Surprisingly on the other hand, for the more challenging ContraPro data, the performance of sentence-level* is comparable to other models.

\begin{table}[t]
\centering
\begin{tabular}{l c c c c}
\hline
& \multicolumn{2}{c}{\textbf{EN$\rightarrow$DE}} & \multicolumn{2}{c}{\textbf{EN$\rightarrow$FR}} \\
\hline
\textbf{Context size} & \textbf{0} & \textbf{5} & \textbf{0} & \textbf{5} \\
\hline
sentence-level & 42  & -- &  76 &  -- \\
sentence-level* &  -- & 47 & --  & 81  \\
concatenation & 45 & 58 & 76 &  85 \\
multi-encoder &  43 & 43 &  76 & 75   \\
\hline
\end{tabular}
\caption{ContraPro contrastive accuracy (\%) for different context sizes. The accuracy is calculated based on the percentage of time a model correctly scores a positive example above its incorrect variant.}
\label{ContraPro-accuracy}
\end{table}


\begin{table}[t]
\centering
\begin{tabularx}{\columnwidth}{l@{~~}c@{~~}c c@{~~}c}
\hline
& \multicolumn{2}{c}{\textbf{EN$\rightarrow$DE}} & \multicolumn{2}{c}{\textbf{EN$\rightarrow$FR}} \\
\hline
\textbf{Model} & \textbf{IWSLT} & \textbf{CPro} & \textbf{IWSLT} & \textbf{CPro}\\
\hline
sentence-level  & 62 & 39 & 70 & 44 \\
sentence-level* & 38 & 45 & 53 & 48 \\
concatenation   & 60 & 48 & 67 & 49 \\
multi-encoder   & 61 & 40 & 70 & 44 \\
\hline
\end{tabularx}
\caption{F1 measure (\%) of models on pronoun resolution phenomena on IWSLT and
ContraPro data. The F1 measure is evaluated based on if a word tagged with a
discourse phenomena in the reference exists and is also tagged in the
hypothesis.}
\label{f-measure-pronouns}
\end{table}

\begin{table*}
\centering
\begin{tabular}{l r@{~}l r@{~}l r@{~}l r@{~}l r r}
\hline
\textbf{Dataset} & 
\multicolumn{2}{c}{\textbf{pronouns}} & 
\multicolumn{2}{c}{\textbf{cohesion}} & 
\multicolumn{2}{c}{\textbf{formality}} & 
\multicolumn{2}{c}{\textbf{verb form}} &
 \textbf{no.\ sent.} & \textbf{no.\ tokens}\\
\hline
\textbf{EN$\rightarrow$ DE} \\
\hline
IWSLT & 180&(0.4\%)	& 569&(1.4\%) &	641&(1.5\%) & --& & 2,271 & 40,877 \\
ContraPro & 14,477&(2.4\%) & 87&(0.01\%) & 9,710&(1.6\%) & --& & 70,718 & 599,197 \\
\hline
\textbf{EN$\rightarrow$FR} \\
\hline
IWSLT & 311&(1.2\%)	& 150&(0.6\%) &	329&(1.3\%) & 787&(3.1\%) &  1,210 & 25,638 \\
ContraPro & 22,810&(2.6\%) & 195&(0.02\%) &  10,505&(1.2\%) & 16,211&(1.8\%) & 81,989 & 865,890 \\
\hline
\end{tabular}
\caption{Discourse phenomena statistics in different datasets along with the total number of the sentences and tokens in each dataset. 
Numbers outside parentheses are counts; numbers inside parentheses indicate percentages of tagged tokens out of the total number of tokens.}
\label{phenomena-statistics}
\end{table*}

\begin{table}[t]
\centering
\begin{tabular}{l c c}
\hline
\textbf{Model} & \textbf{cohesion} & \textbf{formality} \\
\hline
\textbf{IWSLT} \\
\hline
sentence-level  & 68 &	67  \\
sentence-level* & 20 &  29 \\
concatenation   & 67 &  68 \\
multi-encoder   & 66 &  67 \\

\hline
\textbf{ContraPro} \\
\hline
sentence-level  & 29 & 31 \\
sentence-level* & 24 & 33  \\
concatenation   & 27 & 35 \\
multi-encoder   & 31 & 33 \\
\hline
\end{tabular}
\caption{F1 measure (\%) of models on lexical cohesion and formality phenomena on
ContraPro and IWSLT datasets for EN$\rightarrow$DE.}
\label{f-measure-other-phenomena}
\end{table}

\begin{table}
\centering
\begin{tabularx}{\columnwidth}{l c c c}
\hline
\textbf{Model} & \textbf{cohesion} & \textbf{formal} & \textbf{vb.\ form} \\
\hline
\textbf{IWSLT} \\
\hline
sentence-level   & 81 & 71 & 42 \\
sentence-level*  & 36 & 45 & 13 \\
concatenation    & 81 & 75 & 42 \\
multi-encoder    & 82 & 74 & 43 \\
\hline
\textbf{ContraPro} \\
\hline
sentence-level  & 58 & 32 & 28 \\
sentence-level* & 53 & 31 & 26 \\
concatenation   & 56 & 32 & 28 \\
multi-encoder   & 58 & 33 & 29 \\
\hline
\end{tabularx}
\caption{F1 measure (\%) of models on lexical cohesion, formality and verb-form
phenomena on ContraPro and IWSLT datasets for EN$\rightarrow$FR.}
\label{f-measure-fr}
\end{table}

\subsection{Discussion}
Previous sections outlined different evaluation techniques for assessing context utilization of document-level MT models. These evaluations are complementary to each other and equally important. We start with a perturbation analysis to confirm whether the model is utilizing the correct context and it is not just acting as regularization. furthermore, we show that utilizing the correct context is not enough to handle context dependent phenomena; since not all context is important. Therefore, for a more fine-grained evaluation, we assess models in how well they utilize the parts in the context that are necessary to handle the phenomena. For this purpose, we use attribution scores supported with an accuracy evaluation (F1 measure) on the phenomena.

\begin{figure}[t]
    \centering
    \includegraphics[width=\columnwidth]{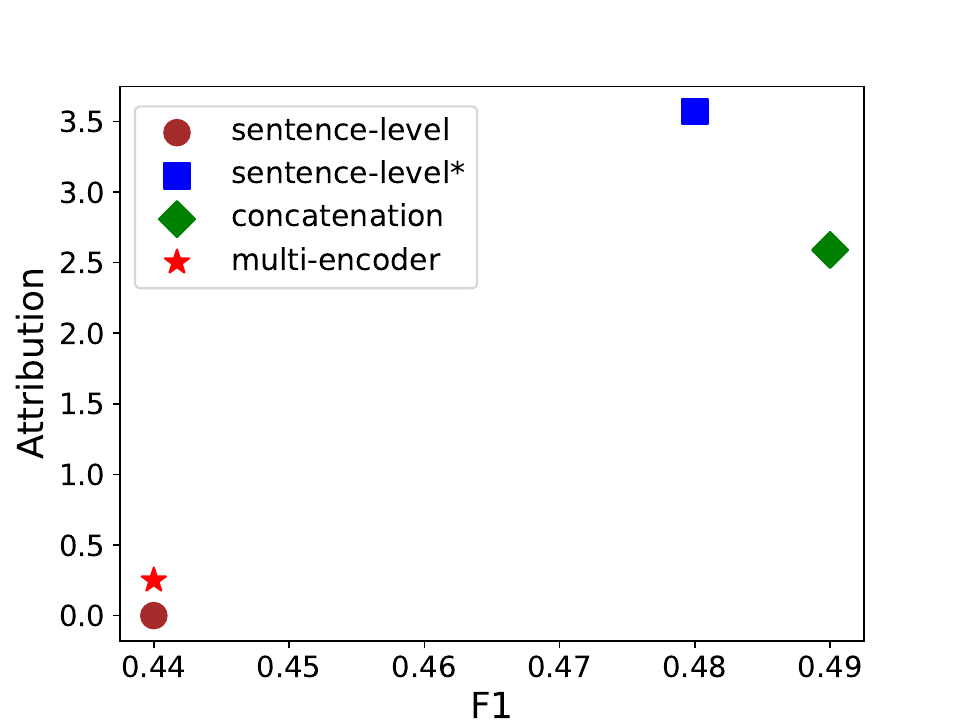}
    \caption{Pareto plot for EN$\rightarrow$FR pronouns. The plot shows that attribution evaluations and accuracy based evaluations are complementary to each other. In particular, there is a trade-off between the sentence-level* and concatenation models, while the multi-encoder and sentence-level models are dominated.}
    \label{pareto-plot}
\end{figure}

Moreover, we show that \textit{supporting context} attribution should be considered as a separate evaluation dimension from pronoun translation quality using Pareto-style plots: \cref{pareto-plot} shows the Pareto plot of two evaluation methods for EN$\rightarrow$FR pronoun resolution: the F1 measure and the supporting context attribution percentage. It can be noticed that the multi-encoder model is sub-optimal on both dimensions, while the sentence-level* and concatenation methods present a trade-off. furthermore, despite the comparable F1 measure of the sentence-level to the multi-encoder model, it has zero attribution.

Overall, our study highlights the important aspects to consider when evaluating context utilization: the use of correct context, the utilization of the correct parts of the context, the accuracy performance on the discourse phenomena, in addition to the general translation performance of course.

\section{Related Work}

    Previous studies on evaluating context influence on MT performance often examined specific context-aware architectures or particular discourse phenomena. \Citet{DBLP:journals/information/NayakHKW22} explored context effects on the hierarchical attention context-aware MT model,  showing that the improved performance on general metrics is due to a context-sensitive class of sentences. \Citet{DBLP:conf/naacl/BawdenSBH18} improved
    the multi-encoder model by encoding the source and context sentences separately while concatenating the current and previous target sentences on the decoder side, demonstrating the importance of target-side context. In contrast, we offer
    a generalizable approach applicable to any context-aware MT model. While we focus on pronoun resolution, our tools can extend to various linguistic phenomena given appropriate rules for annotating \textit{supporting context}.

    In comparing various document-level models, \citet{DBLP:conf/wmt/HuoHGDKN20} found performance variation based on tasks, with no universally superior model. They also highlight back-translation's benefit to document-level systems, noting their robustness against sentence-level noise. Unlike their general metric approach, we enhance the analysis using perturbation methods and attribution evaluation.

    In interpreting context's benefits, \citet{DBLP:conf/discomt/KimTN19} quantified the causes of improvements of context-aware models on general test sets using attention scores. They found that context usually acts as a regularization and is rarely utilized in an interpretable way.  Our work differs in that we use ALTI+ attribution scores instead of attention scores to interpret models' behaviors.

    In a concurrent work, \citet{sarti2023quantifying} introduced
    an end-to-end interpretability pipeline for analyzing context reliance in context-aware models.
    In contrast, we rely on linguistic rules instead of attention weights or gradient norms to extract contextual cues,  which we show to align with human annotated cues. Additionally, we use attribution scores to compare different
    MT models, including single-- and multi-encoder ones.

\section{Conclusion}
In this work, we shed light on multiple angles to look from when evaluating context utilization in document-level MT. We use a perturbation-based analysis to investigate correct context utilization. Additionally, for phenomena-specific evaluation, we propose using attribution scores as measure context utilization. We suggest calculating the attributions of only the supporting context that is necessary for handling context-dependent phenomena. Moreover, we show that automatically annotated supporting context is inline with human annotated supporting context and can be used as an alternative. Finally, we highlight the importance of using discourse-rich data in evaluation.

Based on our proposed analysis and evaluation tools, we argue that
the single encoder approaches to document-level MT demonstrate a priori better context use while also scoring high for translation quality, suggesting that multi-encoder models need more careful design or tuning as highlighted by \citet{DBLP:journals/corr/abs-2109-02995}.

For future work, we aim to extend attribution evaluation to other discourse phenomena, by designing rules for automatic annotation of supporting context for the phenomena with the aid of linguistic expertise. We would also like to apply our evaluation tools and setups to different document-level architectures to provide a solid benchmark of context utilization by context-aware models.

\section*{Limitations}
One limitation is that our conclusions regarding the multi-encoder model are considering only one instance of the multi-encoder approaches to document-level MT. We do not claim that all multi-encoder approaches to document-level MT will have low degrees of context utilization. We leave it to future work to investigate the context utilization of other multi-encoder approaches.

Due to the lack of \textit{supporting context} annotations for discourse phenomena, we focused only on the pronoun resolution phenomena on two language pairs: EN$\rightarrow$DE and EN$\rightarrow$FR. However, we hope that this study encourages more work on automatic \textit{supporting context} annotations for all identified discourse phenomena.

\section*{Broader Impact}
Machine translation is a widely adopted technology relied upon by many people, sometimes in sensitive, high-risk settings such as medical and legal ones \citep{doi:10.1080/1369118X.2020.1776370}.
While here we propose a more multifaceted evaluation of MT systems in hopes of mitigating such risks by identifying less robust systems, our automated evaluation, like any, is imperfect and limited. For systems deployed in critical scenarios, a more bespoke and in-depth analysis is necessary to complement our approach.

\section*{Acknowledgements}
We would like to thank Wilker Aziz, Evgenia Ilia, Pedro Ferreira, Chryssa Zerva, Jose C. De Souza, Catarina Farinha and the LTL team at UvA for their valuable comments and discussions about this work.
This work is part of the UTTER project, supported by the European Union's Horizon Europe research and innovation programme via grant agreement 101070631.
VN also acknowledges support from the Dutch Research Council (NWO) via VI.Veni.212.228.

\bibliography{anthology,custom}

\end{document}